# *Remote Human-Robot Interaction in Greenhouses via Virtual Reality: How Plant Canopy Structure Affects Leaf Disease and Soil Moisture Inspection*

Daniel Udekwe, Hasan Seyyedhasani
School of Plant and Environmental Science, Virginia Tech, Blacksburg VA, USA

Email of corresponding authors
daudekwe@vt.edu (Daniel Udekwe)
seyyedhasani12@vt.edu (Hasan Seyyedhasani)

ABSTRACT
Virtual reality (VR) teleoperation enables remote robotic inspection of plants in greenhouse environments, but inspection performance may be constrained by plant structure and sensor visibility. This study evaluates VR-based remote human–robot interaction for leaf inspection and soil moisture assessment and investigates factors influencing inspection accuracy and task performance. A robotic system comprising an unmanned ground vehicle and a camera-equipped robotic manipulator was controlled through a VR teleoperation interface using kinematic models for navigation and manipulator control. Fourteen plants were evaluated across two experiments. Leaf inspection was assessed using task completion time and disease-detection accuracy, while soil moisture assessment was evaluated based on correct watering decisions. Leaf inspection times ranged from 3.3–8.0 s, with disease detection reaching 88% accuracy; improvement between experiments was not significant ($p = 0.378$). Soil moisture assessment achieved 64.3% accuracy (9/14 plants), with no significant improvement between experiments ($p = 0.50$). However, canopy morphology significantly predicted assessment success ($p < 0.01$), with broad, single-leaf canopies achieving 100% success compared with 16.7% for dense, compound canopies. Operators became faster when inspecting dense canopies without improving accuracy, suggesting that camera occlusion, rather than operator performance, was the primary limitation. These findings demonstrate the importance of accounting for plant morphology in VR-based agricultural robot design and support adaptive camera positioning and multimodal sensing strategies to improve remote greenhouse monitoring reliability.

# 1. INTRODUCTION

The quest for sustainable agriculture has led to an increased interest in leveraging robotics and automation to optimise various aspects of crop cultivation (Fountas et al., 2020; Vrochidou et al., 2021; Udekwe, 2024). In this pursuit, the integration of Human-Robot Interaction (HRI) emerges as a pivotal paradigm, bridging the gap between human expertise and robotic capabilities (Anagnostis et al., 2021; Rysz & Mehta, 2021). HRI facilitates seamless collaboration, enabling robots to perform complex tasks under the guidance of human operators, which enhances the efficiency and precision of agricultural practices (Lytridis et al., 2021; Sparrow & Howard, 2021). This integration is particularly significant in the context of greenhouse agriculture, where meticulous monitoring and precise interventions are paramount.

The synergy between human operators and robotic systems holds immense promise for improving crop yields, reducing labor costs, and minimizing environmental impact (Ju et al., 2022). Studies have demonstrated that robots equipped with advanced sensors and machine learning algorithms can effectively monitor leaf health and soil moisture levels, providing real-time data to inform decision-making processes (Bac et al., 2014; Santos et al., 2020). Moreover, the adaptability of robotic systems to different greenhouse environments further underscores their potential to revolutionise modern agriculture (Bechar & Vigneault, 2016; Shamshiri et al., 2018).

HRI in agriculture has been explored extensively in recent years, emphasizing the importance of integrating advanced robotics with human oversight to enhance the precision and efficiency of agricultural operations (Li et al., 2022). One significant application of HRI is in greenhouse management, where the controlled environment demands accurate and timely interventions to maintain optimal growing conditions (Cailian et al., 2021; Chen et al., 2021; Jiang et al., 2022; Kootstra et al., 2021; Rong et al., 2021; Tiwari et al., 2020). Greenhouses provide a unique setting for deploying robotic systems due to their enclosed nature and the need for constant monitoring of plant health and soil conditions (Akrami et al., 2020; Shamshiri et al., 2018). By employing human-operated robots for tasks such as leaf inspection and soil moisture level observation, human operators can focus on more strategic aspects of crop management, thereby improving overall productivity and sustainability (Wakchaure et al., 2023; Seyyedhasani et al., 2025).

This study explores remote HRI within the specific domain of greenhouse management, focusing on the dual objectives of leaf inspection and soil moisture level assessment. The experimental framework revolves around a novel setup involving a robotic arm mounted on an unmanned ground robot, operated remotely via the internet using virtual reality (VR) interfaces. This innovative configuration, as shown in Figure 1, not only facilitates real-time interaction but also extends the reach of human operators beyond geographical confines, enabling seamless supervision and intervention in greenhouse environments irrespective of physical distance.

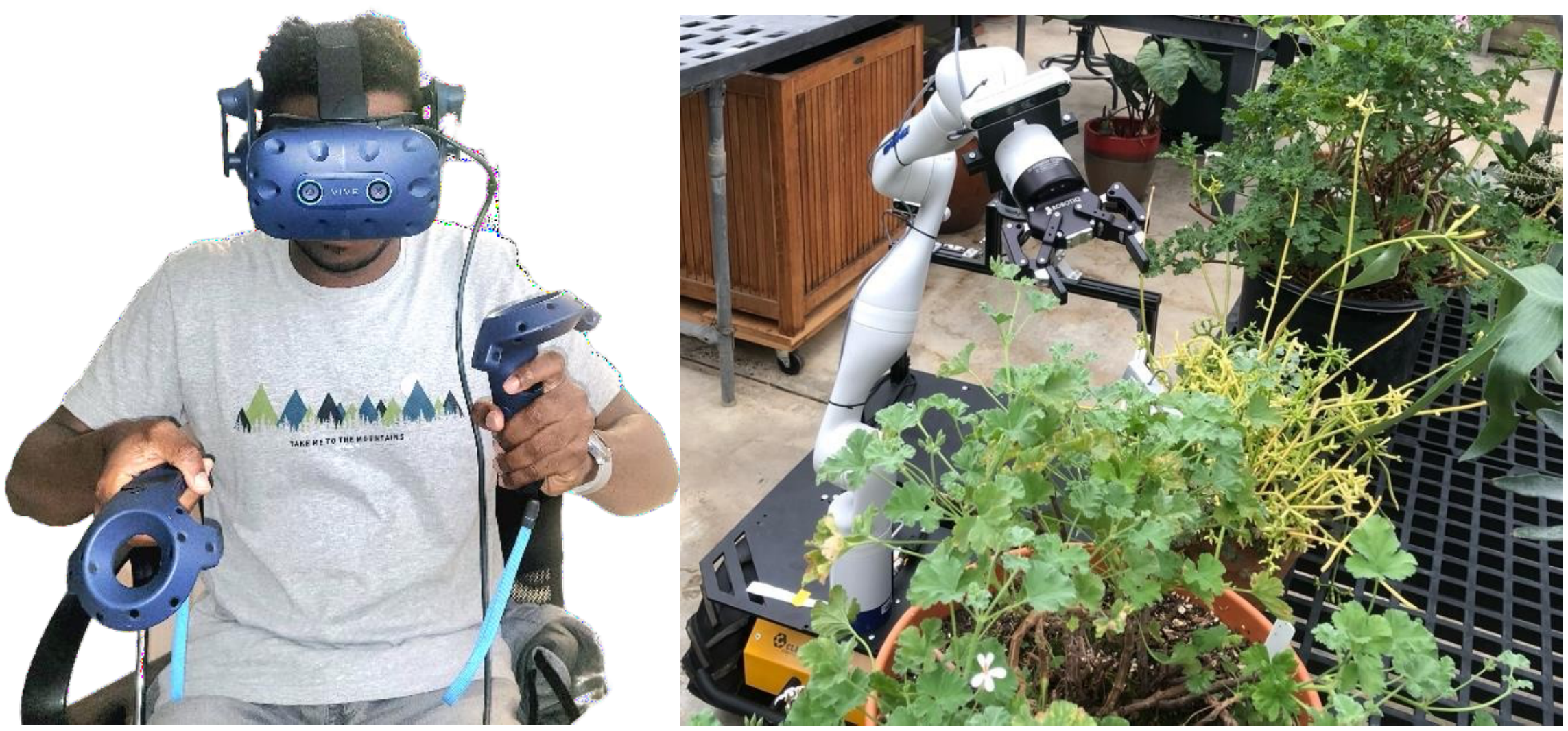

**Figure 1:** A user controlling the robotic platform to inspect a plant for diseases

A robotic manipulator, equipped with high-resolution cameras performs detailed inspections of plant leaves, identifying issues such as pest infestations, nutrient deficiencies, and diseases at an early stage (Mahlein et al., 2024; Santos et al., 2020). Simultaneous observations of data on soil conditions also allows precise irrigation managements. This dual approach ensures that both plant health and soil moisture levels are maintained within optimal ranges, leading to healthier crops and more efficient water usage.

The integration of robotics and automation in agriculture, specifically through HRI, has been a subject of significant research and development. Several studies have highlighted the potential of HRI to transform agricultural practices by enhancing efficiency, precision, and sustainability. For instance, Bac et al. (2014) demonstrated the effectiveness of robots equipped with advanced sensors in monitoring leaf health and soil moisture levels. Their work emphasised the capability of robotic systems to provide real-time data, which is crucial for informed decision-making in crop management.

Remote operation via VR interfaces enhances the user experience by providing an immersive environment where operators can interact with the robotic system as if they were physically present in the greenhouse (Chandana et al., 2020; Opiyo et al., 2021). This technology not only improves the accuracy of interventions but also reduces the need for on-site labor, making it a cost-effective solution for large-scale greenhouse operations. The ability to monitor and manage multiple greenhouses from a central location further amplifies the benefits of this approach, offering a scalable solution to modern agricultural challenges.

The adaptability of robotic systems to different greenhouse environments have been recently studied. Shamshiri et al. (2018)focused on the deployment of robots in controlled environments. Their research indicated that robots could perform precise interventions, thereby maintaining optimal growing conditions and improving crop yields. Bechar & Vigneault (2016) further reinforced this perspective, highlighting the role of HRI in reducing labor costs and minimizing environmental impact through efficient resource management. Opiyo et al. (2021) delved into the use of virtual reality (VR) interfaces for remote operation of robotic systems in agriculture. Their findings suggested that VR technology enhances user experience and accuracy of interventions, making it a viable solution for large-scale operations.

Overall, these studies collectively underscore the transformative potential of HRI in agriculture, paving the way for innovative solutions that address modern agricultural challenges. The integration of advanced robotics with human oversight not only enhances the precision of agricultural practices but also offers scalable solutions for sustainable farming. The primary contribution of this paper is an empirical characterization of how plant canopy morphology governs the reliability of camera-based remote sensing tasks in VR-teleoperated greenhouse HRI. Through a two-experiment case study spanning 14 morphologically diverse plants, this work investigates whether soil moisture assessment success is primarily limited by operator skill and system familiarity, or by whether the plant's canopy structure permits an unobstructed camera line of sight to the target surface. In doing so, this study aims to identify whether occlusion, rather than control precision or interface latency, constitutes the dominant engineering constraint for this class of system, with a view toward informing concrete design implications for future greenhouse HRI platforms regarding camera viewpoint strategy and sensing modality. As a secondary aim, this study examines the relationship between cycle completion time and task success across plant morphology classes, to assess whether occlusion-driven failure is decoupled from operator speed and effort and is therefore better characterized as a sensing limitation rather than a control or training deficiency.

Building on this motivation, this study addresses the following research questions:

- RQ1: How accurately and efficiently can a VR-teleoperated dual-robot system (an unmanned ground vehicle paired with a robotic arm) perform leaf disease inspection and soil moisture assessment across morphologically diverse greenhouse plants?
- RQ2: Does plant canopy morphology influence the reliability of camera-based remote sensing tasks performed through VR teleoperation, and if so, through what mechanism?
- RQ3: How does operator experience across repeated trials affect teleoperation performance for these two tasks?

Correspondingly, this study tests the following hypotheses:

- H1: The teleoperated robotic system will achieve measurable and repeatable cycle completion times and diagnostic/assessment success rates for both leaf disease inspection and soil moisture assessment.

- H2: Plants with dense, compound canopy structures will exhibit lower soil moisture assessment success rates than plants with broad, single-leaf canopies, due to camera occlusion of the target soil surface.
- H3: Operator performance, as measured by cycle completion time and success rate, will improve between the first and second experimental trials as a result of increased familiarity with the VR teleoperation interface.
- H4: For occlusion-affected (dense-canopy) plants, task success will remain low regardless of reductions in cycle completion time, indicating that occlusion imposes a sensing limitation that operator speed or effort cannot overcome.

The structure of the remaining of this paper is outlined as such: Section 2 provides a description of all major components of the system as well as the methodology employed. The results obtained from the disease inspection and moisture observation experiments are provided and discussed in Section 3 and the conclusion is given in Section 4.

# 2. MATERIALS AND METHODS

## 2.1 System Design

An overview of the proposed system is shown in Figure 2. This figure illustrates the interconnection of the system components for the teleoperated robotic system.

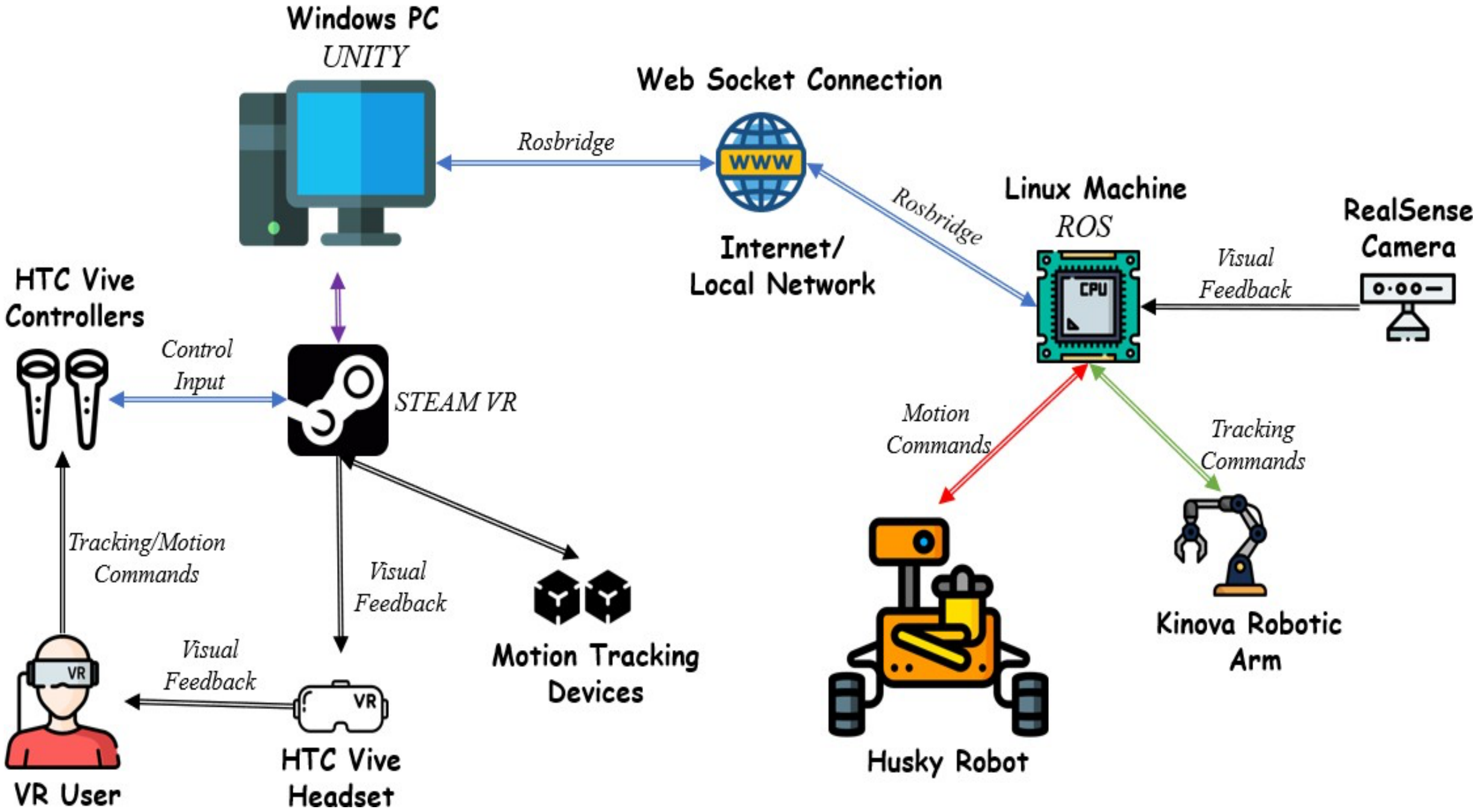

**Figure 2:** The complete framework of the developed VR-based teleoperation system (Udekwe & Seyyedhasani, 2025a, 2025b)

The specifics of the designed system in Figure 2 and its capabilities are outlined by Udekwe & Seyyedhasani (2025a). The following sections provide detailed descriptions of the operation of these components that form the VR subspace and the remote environment.

## 2.2 VR Subspace

The VR subspace comprises the user, the windows PC running Unity as well as the VR headset and hand controllers as shown in Figure 3. The Unity serves as the central platform for creating the graphical user interface (GUI) and the immersive virtual reality (VR) environment. It facilitates the visualization of data collected by the robot and enables user interaction with the virtual representation of the greenhouse. The VR subspace includes camera feeds from the gripper and front-mounted cameras, as well as a 3D replica of the robots to visualise the robot's pose at all points in time. This provides an immersive viewing experience, allowing the human operator to visualise the greenhouse environment from the robot's perspective. Hand controllers enable intuitive interaction with virtual objects in the VR space, empowering users to manipulate the robot's movements and perform inspection tasks with precision.

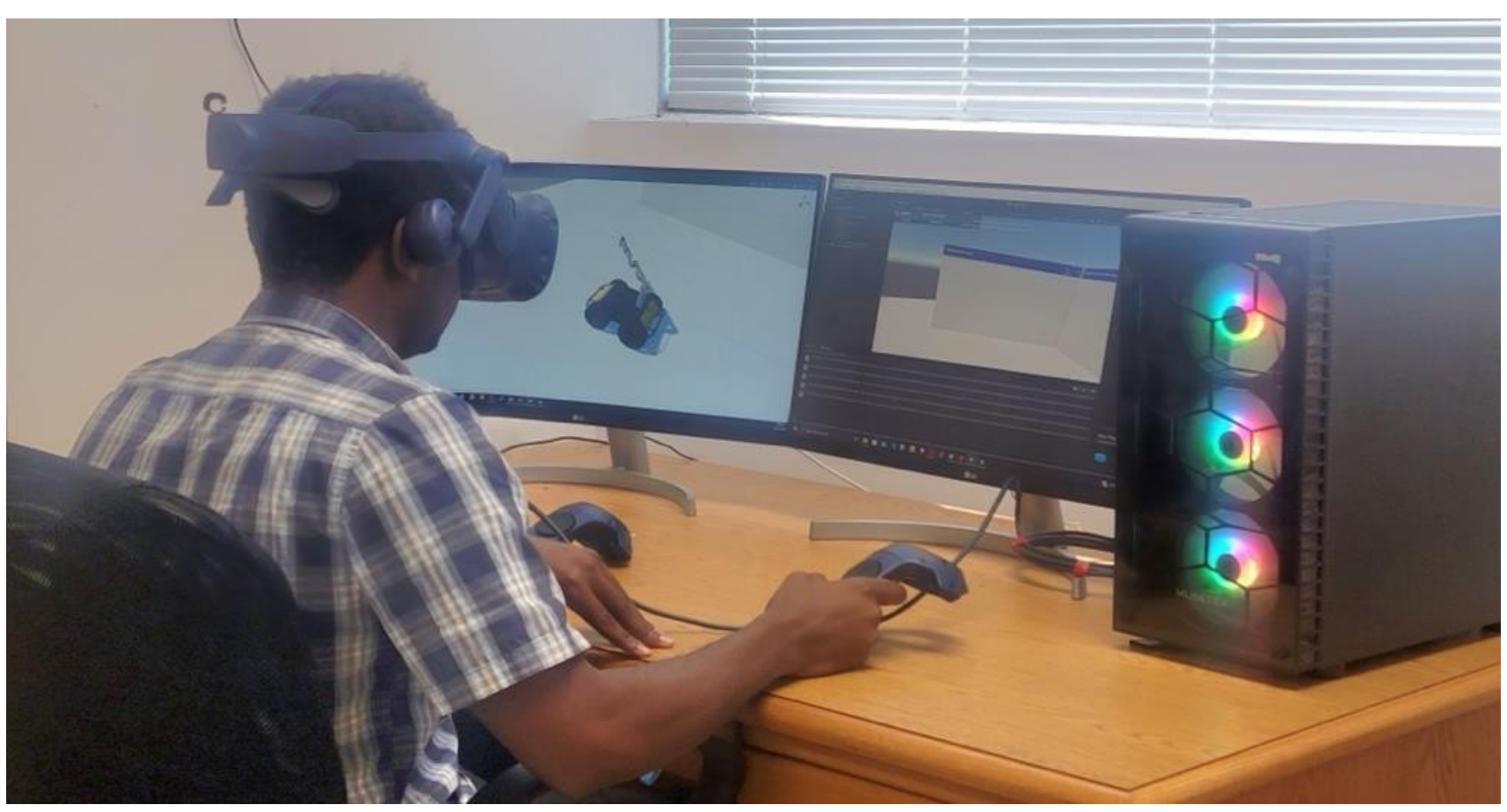

**Figure 3:** A user interacting with the proposed system using the headset and controllers on Unity 3D

### *2.2.1 Motion stabilization*

A single exponential filter was used in the VR hand controllers to smooth and stabilise motion data. This filter reduces the jitter and noise inherent in raw sensor data, providing a more fluid and natural user experience. It operates by applying a weighted average to the current and previous data points, with the weighting factor determining the level of smoothing. The result is a more stable and precise tracking of hand movements, essential for immersive and responsive VR interactions.

The single exponential filter can be represented mathematically as (Barrow et al., 2020):

$$S_t = \alpha y_{t-1} + (1-\alpha)S_{t-1}, \quad (1)$$

Where $S_t$represents the smoothed observation, $y$is the original observation at time interval of $t$, and $\alpha$is a smoothing constant within the bounds $0 < \alpha \leq 1$.

Expanding equation (1) and substituting for $S_{t-1}$yields:

$$S_t = \alpha y_{t-1} + (1-\alpha)[\alpha y_{t-2} + (1-\alpha)S_{t-2}], \quad (2)$$

$$S_t = \alpha y_{t-1} + \alpha(1-\alpha)y_{t-2} + (1-\alpha)^2 S_{t-2}, \quad (3)$$

By substituting recursively for $S_{t-2}$, equation (3) can be rewritten as:

$$S_t = \alpha \sum_{i=1}^{t-2} (1-\alpha)^{i-1} y_{t-i} + (1-\alpha)^{t-2} S_2 \; t \geq 2, \tag{4}$$

Equation (4) represents a weighted average of the previous smoothed velocity ($S_{t-1}$) and the observed velocity at the previous time step ($Y_{t-1}$). The weight assigned to $Y_{t-1}$is determined by the smoothing factor $\alpha$. When $\alpha$is close to 1, the most recent observation is given more weight, causing the smoothing to respond quickly to changes. Conversely, when $\alpha$is close to 0, more weight is given to the previous smoothed velocity, resulting in a slower response to changes.

Figure 4 shows a plot of this smoothing algorithm applied to the X values of the right controller. With a smoothing factor $\alpha = 0.02$, the smoothed values are less sensitive to random hand movements along the X axis. Additionally, Figure 5 displays the tracking of the end effector link of the robotic arm during one of the experiments.

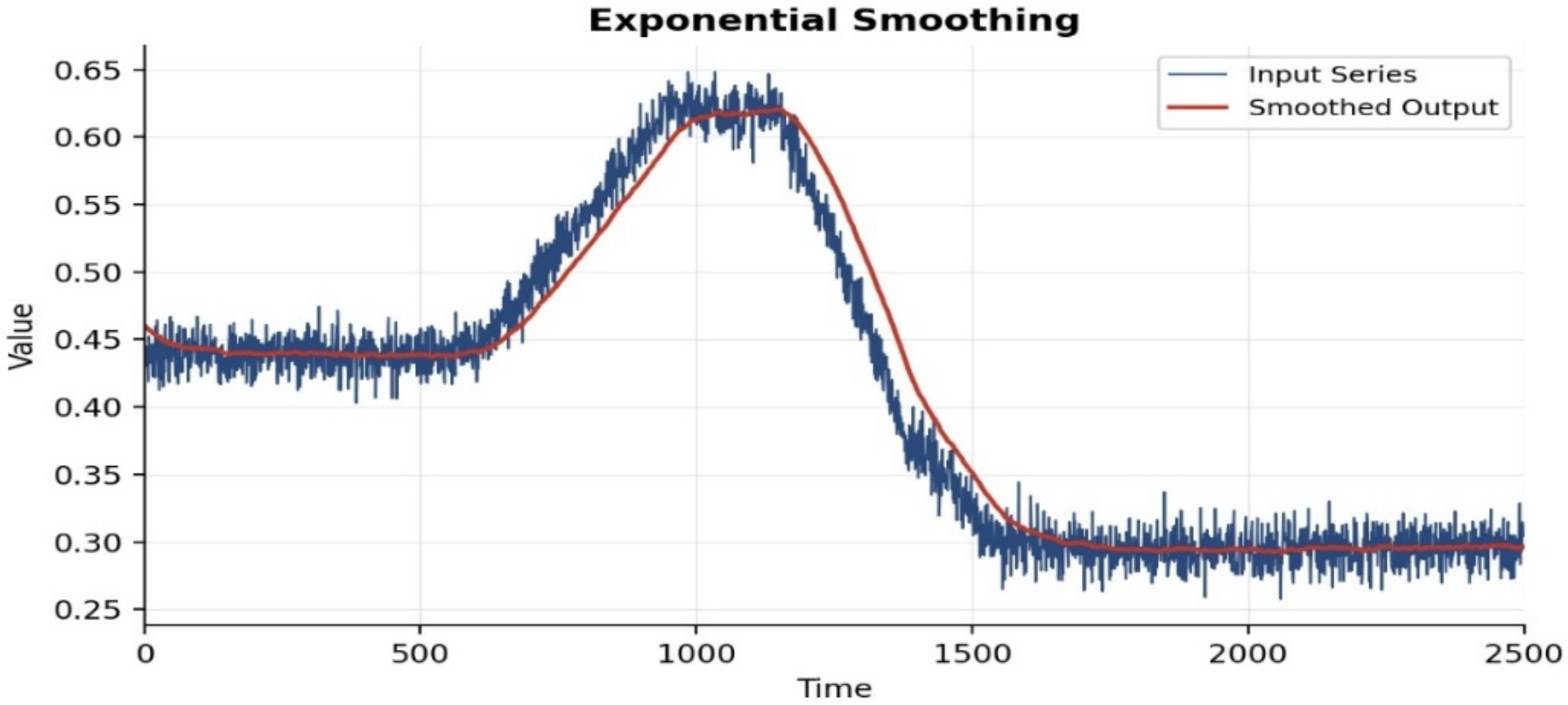

**Figure 4:** A plot showing the single exponential smoothing applied to the X-values of the right controller

**Figure 5:** Position tracking of the end effector link along the x, y and z axis of the robotic arm during one of the experiments.

## 2.3 Remote Environment

The mobile robotic platform is the central component of the remote environment. At the core of this platform is the Robot Operating System (ROS), which acts as its backbone. The Husky Unmanned Ground Vehicle (UGV) manages autonomous navigation within the greenhouse, while the Kinova Gen3 robotic arm is responsible for precise tasks such as leaf inspection and soil moisture assessment. Integrated with the ROS, a Realsense camera is mounted on the platform to capture high-resolution images of plant leaves and soil surfaces. These images are processed to extract crucial features like leaf health indicators and soil moisture levels, which are then displayed to the user via the Unity interface. To facilitate communication, a WebSocket connection is established between the Unity application and the ROS-based robots. This connection ensures real-time, bidirectional communication, allowing for the seamless exchange of control commands, sensor data, and status updates between the human operator interface and the robotic system.

### *2.3.1 Kinematic Modeling of the Mobile Manipulator System*

The Husky UGV and Kinova Gen3 arm are governed by well-established kinematic relations that map operator commands, received via the VR interface, to physical robot motion. This section presents the kinematic models underlying the navigation and manipulation behavior described above.

The Husky UGV is a differential-drive mobile platform: its motion is controlled by independently commanding the angular velocities of its left and right wheels, $\omega_l$ and $\omega_r$. The platform's linear velocity v and angular velocity $\omega$ about its vertical axis are related to the wheel velocities by the wheel radius r and the wheelbase $L$ as:

$$v = \frac{r}{2}(\omega_R + \omega_L), \qquad \omega = \frac{r}{L}(\omega_R - \omega_L) \tag{5}$$

Given $v$ and $\omega$, the UGV's pose $(x, y, \theta)$ in the world frame evolves according to the standard unicycle kinematic model:

$$\dot{x} = v cos\theta, \qquad \dot{y} = v sin\theta, \qquad \dot{\theta} = \omega \tag{6}$$

The Kinova Gen3 arm's forward kinematics are obtained using the Denavit–Hartenberg (DH) convention. Each link $i$ is described by a homogeneous transformation relative to link $i - 1$, parameterized by joint angle $\theta_i$, link offset $d_i$, link length $a_i$, and link twist $\alpha_i$:

$$T_i^{i-1} = Rot_z(\theta_i)\, Trans_z(d_i)\, Trans_x(a_i)\, Rot_x(\alpha_i) \tag{7}$$

Composing these single-link transformations from the base frame to the end-effector frame (link $n$) yields the arm's overall forward kinematics:

$$T_n^0 = \prod_{i=1}^{n} T_i^{i-1} = T_1^0 T_2^1 \ldots T_n^{n-1} \tag{8}$$

For velocity-level control during teleoperation, the relationship between joint velocities $\dot{q}$ and end-effector linear and angular velocity $\dot{x_e}$ is given by the manipulator Jacobian $J(q)$:

$$\dot{x_e} = j(q)\dot{q}, \qquad j(q) \in \mathbb{R}^{6\times n} \tag{9}$$

This Jacobian relation is what allows operator hand velocities captured by the VR controllers to be converted into joint commands for the Kinova arm in real time. Within the closed-loop architecture shown in Figures 6 and 7, the pose tracking error $e(t)$ that drives the robot controller is defined as the difference between the desired pose $x_d(t)$, set by the operator's VR controller input, and the actual measured pose x_a(t) of the robot end-effector or platform:

$$e(t) = x_d(t) - x_a(t) \quad (10)$$

This error signal is continuously fed back to the robot controller, which adjusts the commanded joint and wheel velocities to drive e(t) toward zero, forming the basis of the real-time teleoperation control loop described in Section 2.3.

Figure 6 illustrates the control loop for the local and remote workspaces, where $\{x, y, z\}$ and $\{\phi, \theta, \psi\}$ denote the positions and orientations of the controller and robot, respectively. The subscript "d" signifies the desired position/orientation of the robot, while "$a$" indicates the actual position/orientation. Additionally, Figure 7 presents the closed-loop representation of the entire system.

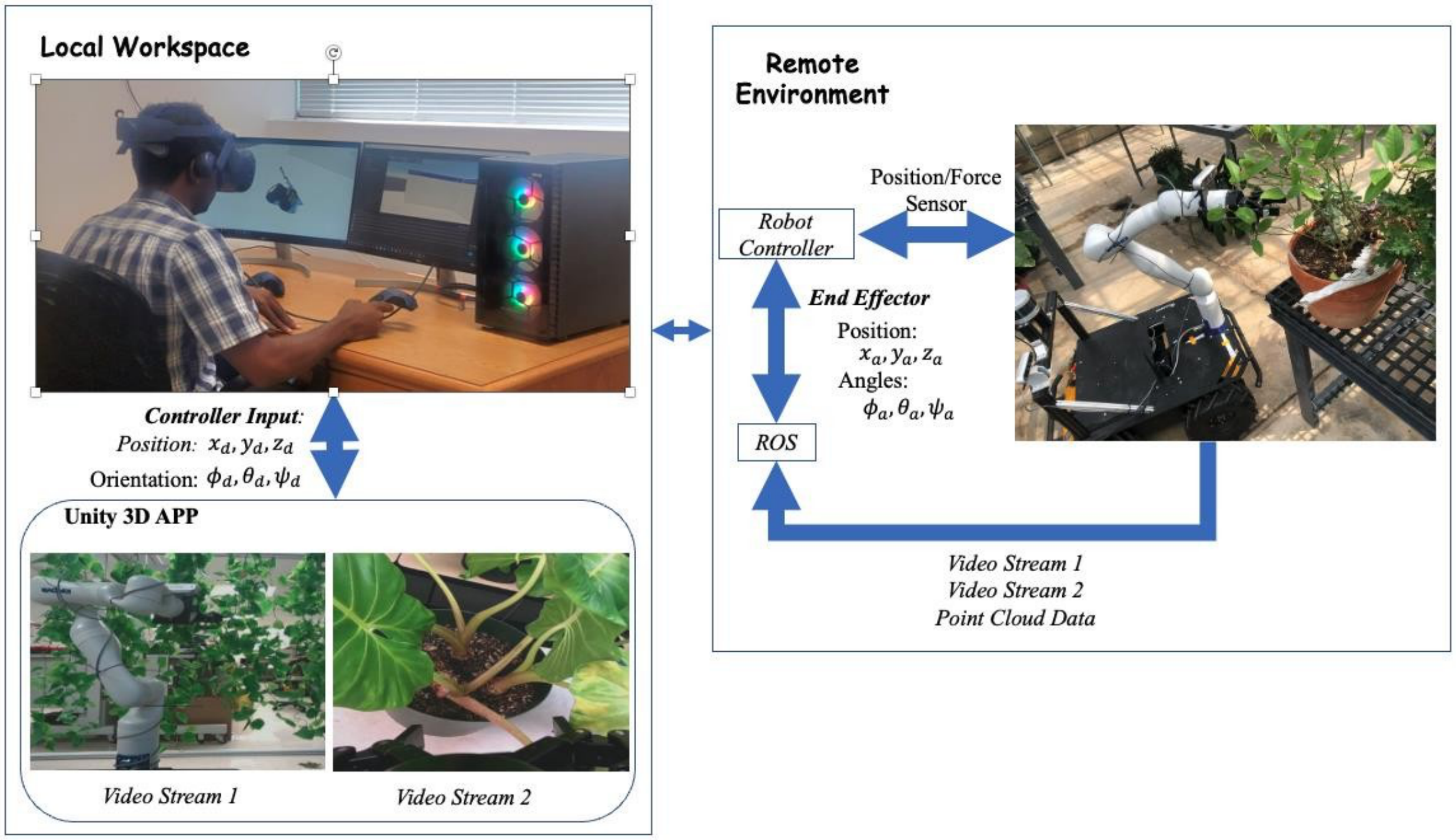

**Figure 6:** The architecture of the VR-based teleoperation system

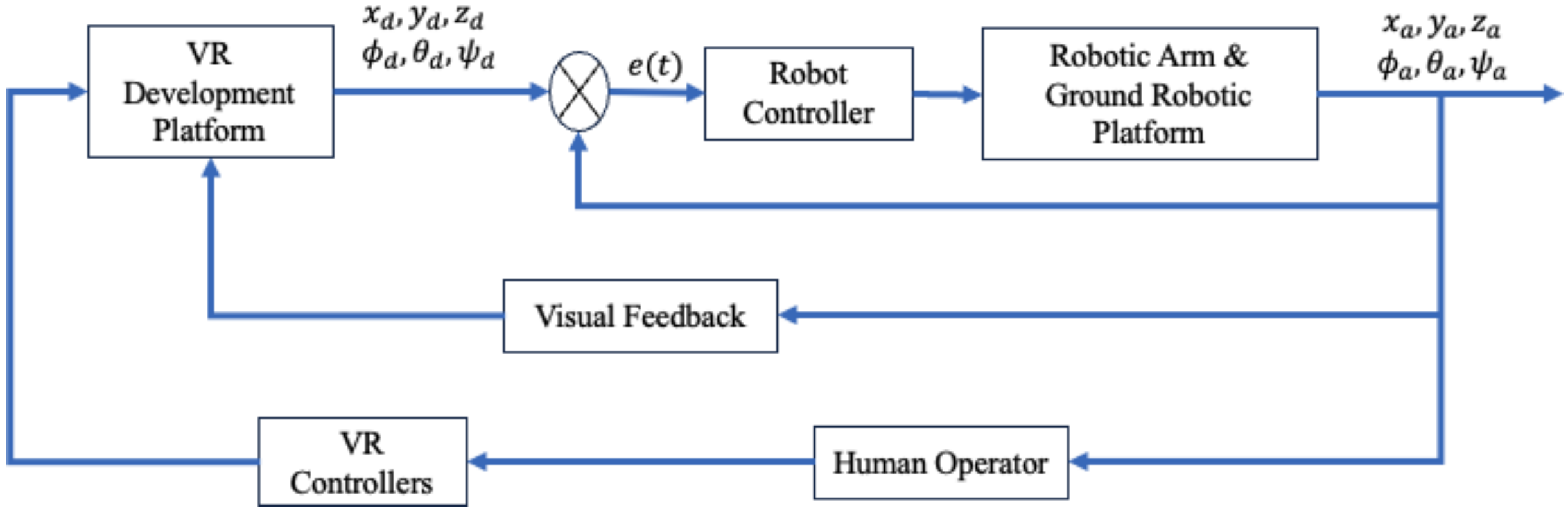

**Figure 7:** Control loop representation of the overall system

The control process starts with a human operator using VR controllers, such as those from the HTC Vive, to send input signals. These signals are processed by a VR development platform, like Unity 3D, which converts the operator's actions into digital commands. These commands are then transmitted to the robot controller, which oversees the operation of both the robotic arm, such as the Kinova Gen3, and the ground robotic platform, such as the Husky UGV. The robot controller executes these commands, coordinating the movements of the robotic arm and UGV to ensure synchronised actions as intended by the operator. Feedback from sensors in both the robotic arm and UGV is sent back to the operator via the VR system, closing the loop and enabling real-time adjustments based on the error signal $e(t)$ defined in Equation (10). This signal represents the discrepancy between the commanded and actual positions and orientations of the arm. This iterative process creates a seamless interaction loop between the human operator, the VR platform, and the robotic components, ensuring precise and responsive control over the entire system.

Figure 8 summarizes the complete operational sequence of the system described in the preceding sections, from initialization through the completion of both experiments for a given plant set.

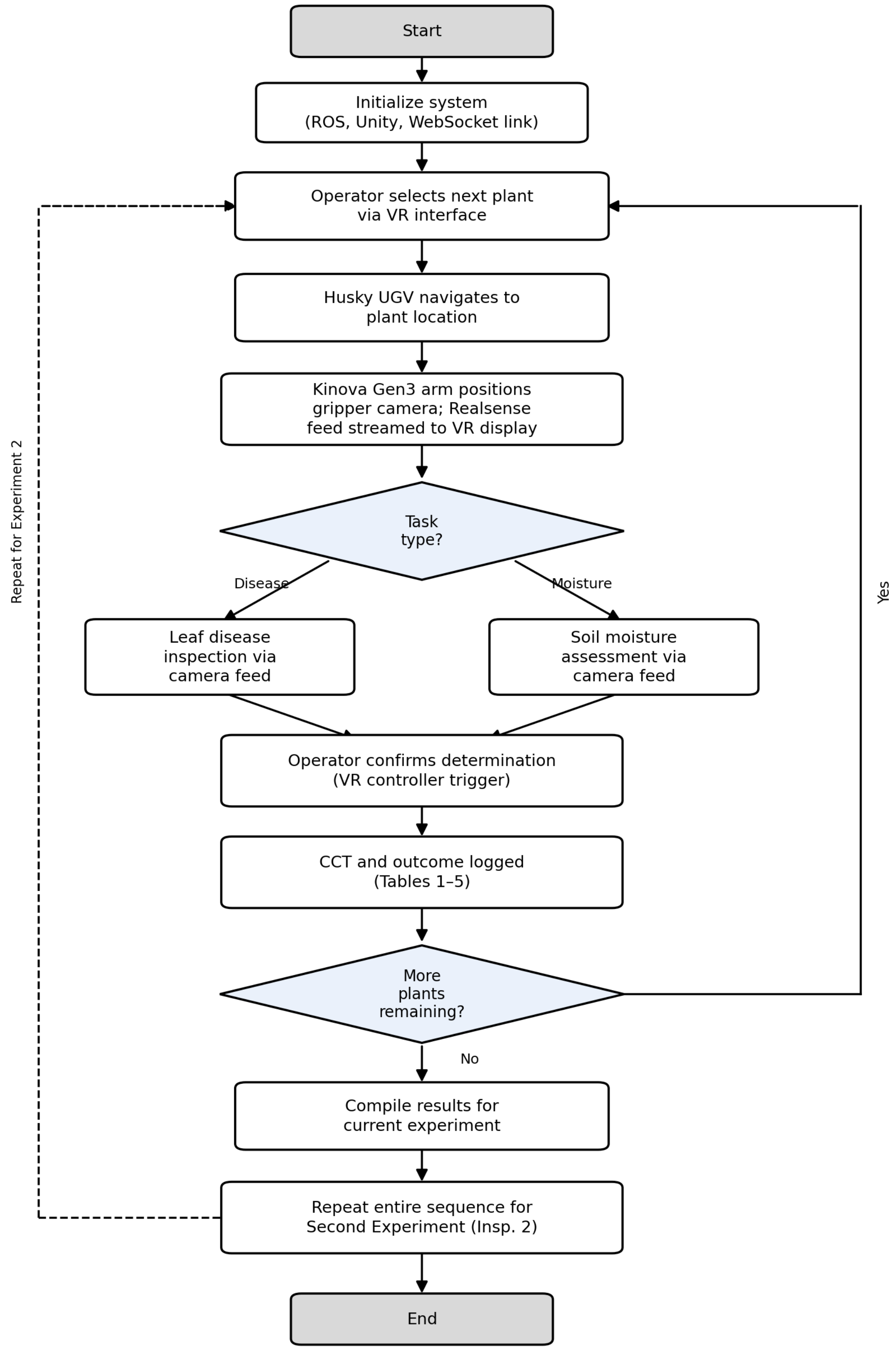

**Figure 8:** Flowchart of the overall system operation, from initialization through both experiments

## 2.4 Experiment Design

To evaluate the effectiveness of the vision-based human-robot-plant interaction system for greenhouse gardening, discussions were held with two gardening experts in the greenhouse facilities. The greenhouses are located on the campus of Virginia Tech, Blacksburg, VA dedicated for research and teaching (Figure 9).

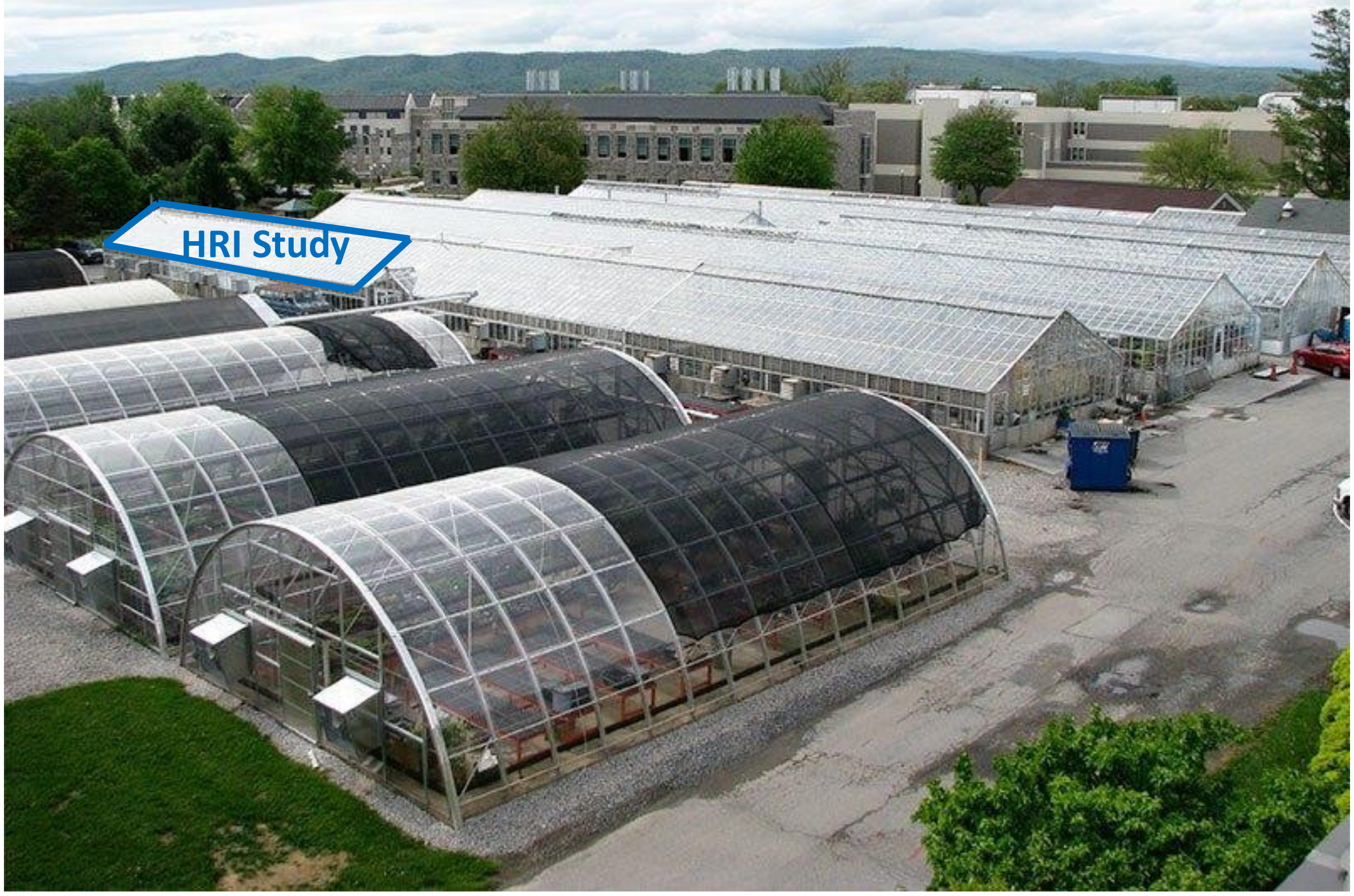

**Figure 9.** The greenhouse facilities located on the campus of Virginia Tech dedicated for research and teaching. The blue box indicates the greenhouse used in the study.

According to the suggestions of the experts, two different experiments were designed for greenhouse gardening: (1) leaf disease inspection and (2) soil moisture level assessment. In the first experiment, the teleoperated robotic system was employed to inspect the leaves of plants for signs of disease. The second experiment involved using the robotic system to assess the moisture levels in the soil. The purpose of utilizing this teleoperated robotic system was to combine the precision of robotic technology with the expertise of human operators, thereby enhancing the efficiency and effectiveness of agricultural management practices. By employing teleoperation, operators can remotely control the robotic system to navigate through the greenhouse environment and conduct inspections with greater flexibility and convenience. This approach enables real-time monitoring of both leaf health and soil moisture levels, allowing for prompt detection and response to potential issues such as disease outbreaks or irrigation needs.

Plants in 14 distinct pots were examined. In case 1, there was a zonal geranium plant. Case 2 contained a variety of plants. Case 3 featured a taro plant, while case 4 had a rose geranium. Case 5 included a little warty, and case 6 had a rose. Case 7 contained a konjac, and case 8 featured a flaming flower. Case 9 housed a Gollum Jade, case 10 a Giant Taro, and case 11 a common coleus. Cases 12, 13, and 14 contained rose geranium, giant taro, and rose geranium, respectively. Images of a few of these plants are shown in Figure 10.

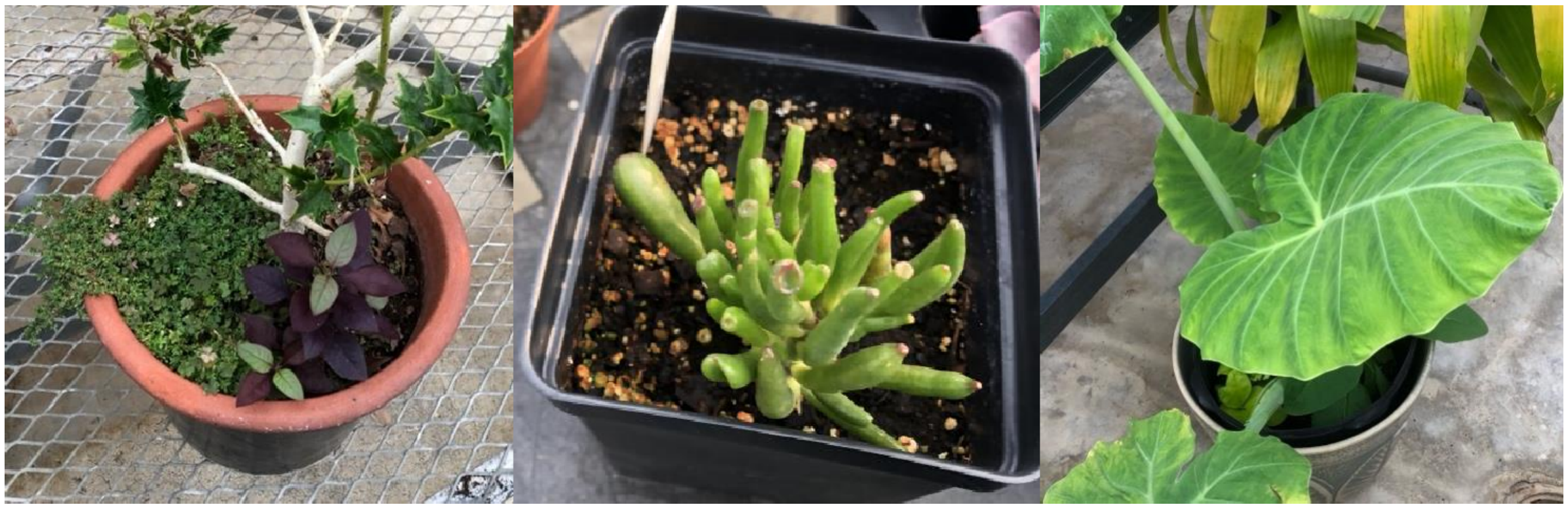

**Figure 10:** Random selection of plants used in the greenhouse experiments

The teleoperation trials for both experiments were conducted by a single operator. As a result, improvements observed between Experiment 1 and Experiment 2 reflect a combination of operator learning and system performance that cannot be fully disentangled with the current single-operator design; this limitation is discussed further in Section 4.

*2.4.1 Leaf Disease Inspection*

This experiment involved navigating the unmanned robotic platform through the greenhouse and manipulating the arm to inspect several of the selected plants. This experiment was performed twice and its efficacy and efficiency evaluated in terms of cycle completion time (CCT) per inspection as well as success rate in correctly detecting a leaf disease which was ascertained by a human expert. Through this experiment, the aim was to evaluate the possibility of substituting the labor-intensive and tedious task of disease detection in a greenhouse with a robotic system.

*2.4.2 Soil Moisture Level Assessment*

The second experiment involved inspecting the moisture levels of each pot soil using the VR system. As with the leaf disease inspection, this experiment involved navigating the robotic platform through the greenhouse and manipulating the robotic arm to inspect the soils for moisture content. The success of the experiment was determined by the CCT of the assessment and the success rate of the VR system verified by a human expert. This experiment's aim was to determine the possibility of deploying the proposed system as a substitute for human operation in a greenhouse setting for moisture level assessment.

*2.4.3 Ground Truth Verification*

To establish ground truth for the leaf disease inspection experiment, a single human expert visually assessed each of the 14 plants in person prior to the teleoperation trials, recording the number and location of visible diseased spots per plant. As this is a proof-of-concept study conducted with a single evaluator, inter-rater reliability was not assessed; the expert was not blinded to which plants had been flagged during system operation, which is a limitation discussed further in Section 4.
For the soil moisture assessment, ground truth watering need was established using a human expert's tactile and visual assessment of the soil surface. As this judgment-based ground truth is inherently more subjective than an instrument-based reading (e.g., a soil moisture meter), it represents a limitation of the current protocol and is discussed further in Section 4.

*2.4.4 Performance Metrics*

To assess the effectiveness of the proposed system, the focus was on two key metrics: cycle completion time (CCT) and success rate of robot-plant interaction (RPI). These metrics directly impact the system's performance. The CCT quantifies the duration required for users to remotely guide the mobile robotic platform (comprising a mobile robot with a robotic arm) to designated points of interest and execute necessary manipulations during experiments. In the leaf disease inspection scenario, the CCT measures the time, in seconds, taken to inspect each plant for diseases. Similarly, in the soil moisture assessment experiment, the CCT indicates the time needed to observe and determine if plants require watering. The RPI success rate assesses the robot's efficiency in accurately identifying diseases and evaluating soil moisture levels by interacting with plants, benchmarked against human expert performance. CCT was measured from the moment the operator-initiated navigation toward a target plant to the moment the operator pressed the VR hand-controller trigger button to confirm a final disease or moisture determination.

For the disease inspection experiment, success was scored at two levels: spot-level success, where an operator-identified diseased spot matched a ground-truth diseased spot on the same leaf; and plant-level success, where at least one diseased spot was correctly identified on a plant confirmed to have disease. For the soil moisture assessment, a trial was scored Pass if the operator's watering-needed determination matched the ground truth established in Section 2.4.3 and Fail otherwise.

The number of plants (n = 14) was constrained by greenhouse capacity and the availability of disease-affected specimens at the time of this proof-of-concept study, rather than a formal power calculation; as several per-plant success rates are computed on small denominators (e.g., a single diseased spot per plant), these rates should be interpreted as descriptive rather than statistically stable estimates. Differences in CCT between Experiment 1 and Experiment 2 were evaluated using a paired t-test, and differences in pass/fail success rates were evaluated using McNemar's test for paired binary outcomes.

# 3. RESULTS AND DISCUSSION

This section provides an explanation and analysis into the experimental results obtained from the leaf disease inspection and soil moisture level assessment experiments.

## 3.1 Plant Disease Inspection

An image captured by the camera affixed to the gripper while inspecting a Flamingo flower is shown in Figure 11. As shown, it is evident that the vision system is capable of spotting unhealthy patches on the leaves. These patches due to malnourishment or disease infestation pose a threat to the overall plant health.

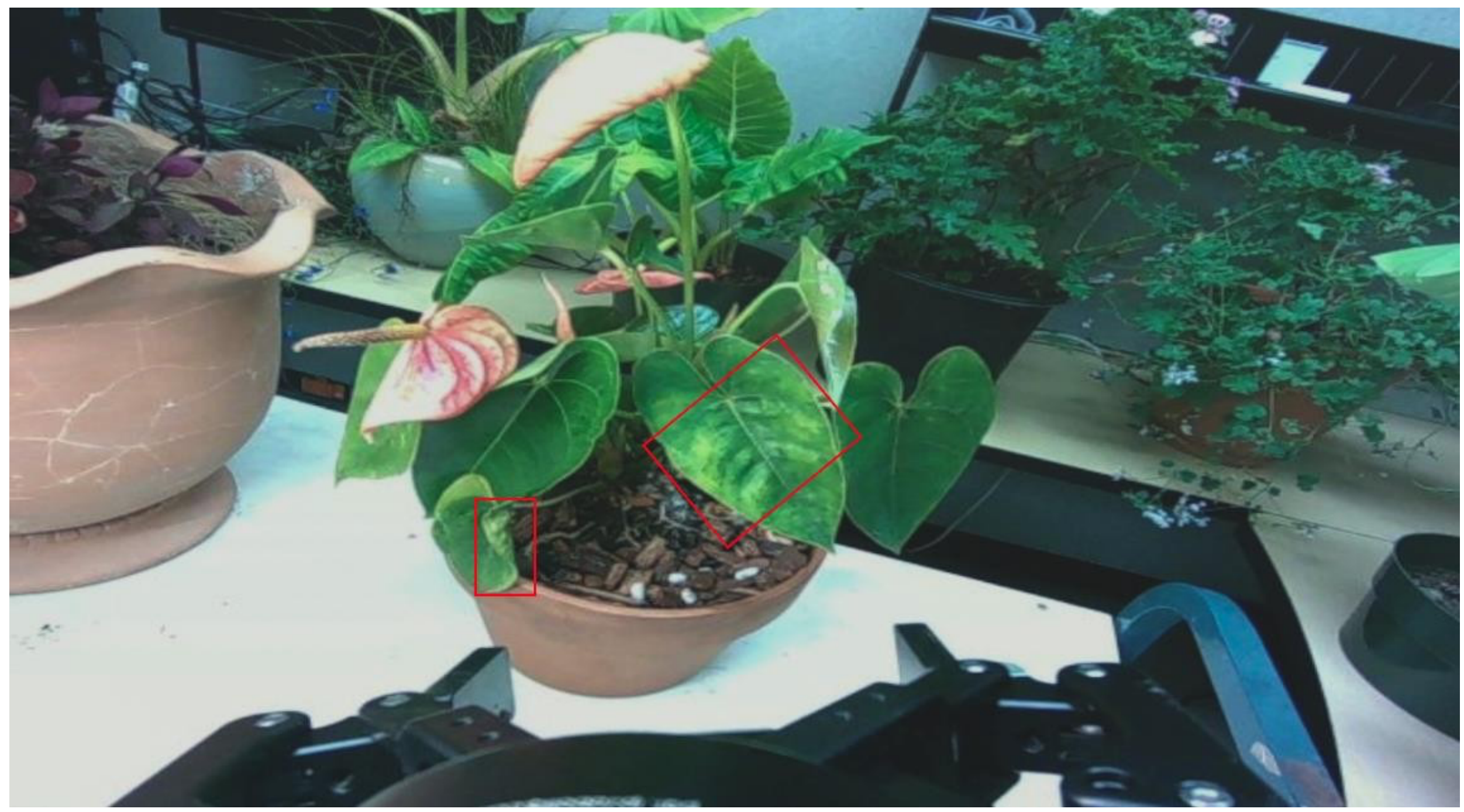

**Figure 11:** An image of a Flamingo flower captured by the gripper camera during the plant disease inspection experiment depicting two diseased spots

The cycle completion time for the plant disease inspection experiment is presented in Figure 12; and the success rate of the robotic system for this experiment is given in Table 2.

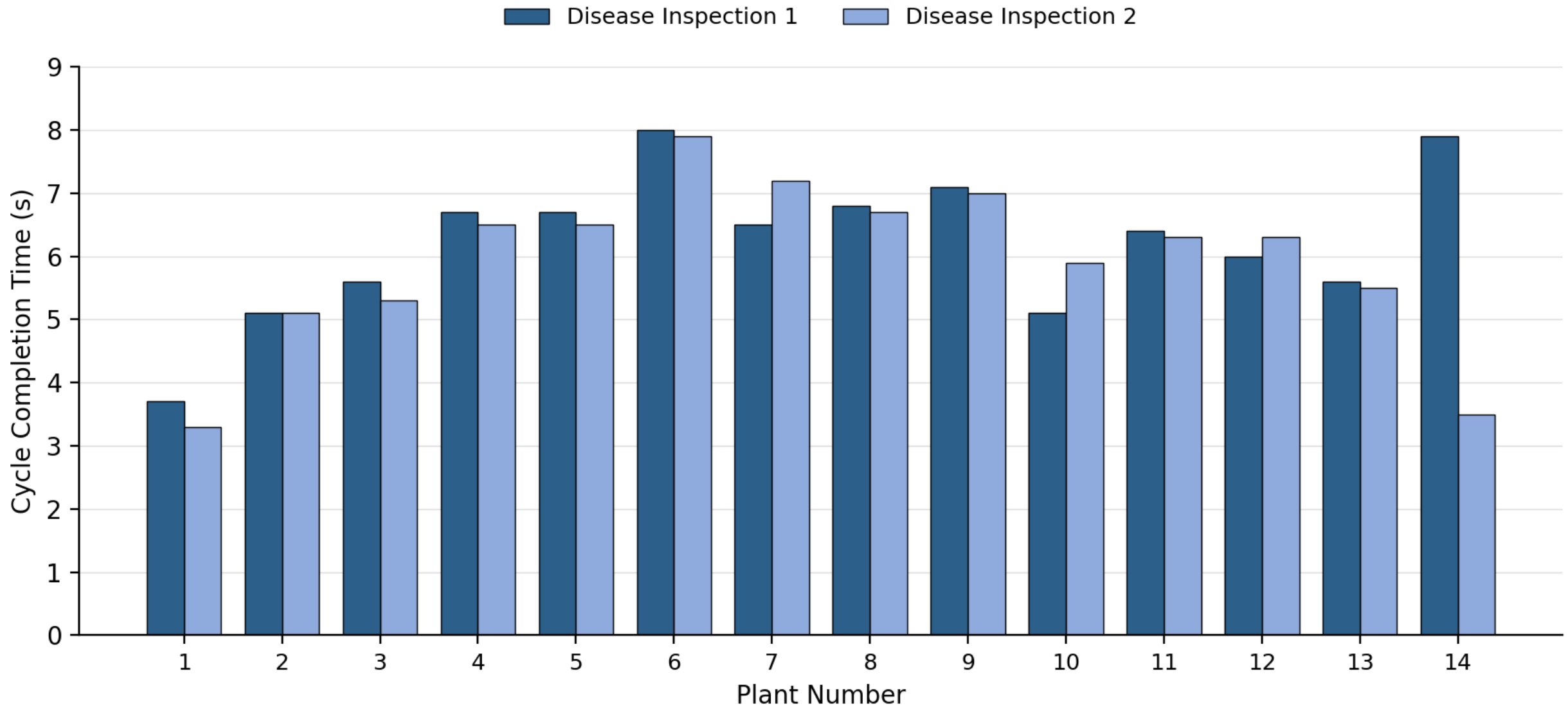

**Figure 12:** Completion times across the disease inspection experiments

**Table 1.** Descriptive statistics of the CCT for the disease inspection across 14 heterogenous plants

| Experiment | # of inspected plants | Min (s) | Max (s) | Mean (s) | STD (s) |
|---|---|---|---|---|---|
| Disease_Insp_1 | 14 | 3.7 | 8.0 | 6.2 | 1.2 |

| Disease_Insp_2 | 14 | 3.3 | 7.9 | 6.0 | 1.3 |
|---|---|---|---|---|---|

Figure 12 represents the cycle completion times (in seconds) for inspecting 14 distinct plants using the proposed system and the summary of these statistics is provided in Table 1. In the first experiment, the cycle completion times varied greatly, ranging from 3.7 s to 8.0 s. The initial plant inspection time was notably the fastest at 3.7 s, while other plants, such as the sixth and fourteenth, had longer inspection times of 8.0 and 7.9 s, respectively. This variation could be due to the complexity or density of the foliage, which could make it more challenging for the robotic arm to navigate and inspect the leaves thoroughly. Additionally, variations in plant size, shape, and health could affect the inspection duration, as plants with more extensive or intricate leaf structures might require more time to inspect. Environmental conditions within the greenhouse, such as lighting or obstacles, could also play a role in extending the inspection times for specific plants.

In the second experiment, there was a slight numerical reduction in the average cycle completion times. The fastest time recorded was 3.3 s, again during the initial plant inspection, whereas the longest time reduced to 7.9 s for the sixth plant. Several plants showed decreased inspection times in the second experiment, such as the eighth and ninth plants, which dropped from 6.8 to 6.7 and 7.1 to 7.0 s, respectively. However, a paired t-test comparing CCT across all 14 plants between the two experiments did not find this reduction to be statistically significant ($t=0.91$, $p=0.378$), so this pattern should be interpreted as a modest, non-significant trend rather than confirmed evidence of operator learning or system improvement between trials.

**Table 2:** Success rates of robot-plant interaction in remote leaf disease detection

| Plant ID | 1 | 2 | 3 | 4 | 5 | 6 | 7 | 8 | 9 | 10 | 11 | 12 | 13 | 14 |
|---|---|---|---|---|---|---|---|---|---|---|---|---|---|---|
| Disease_Insp_1 | 2/2 | 0/0 | 3/5 | 0/0 | 0/0 | 0/1 | 0/0 | 2/2 | 0/0 | 1/3 | 0/0 | 1/2 | 0/2 | 1/3 |
| Disease_Insp_2 | 2/2 | 0/0 | 3/5 | 0/0 | 0/0 | 1/1 | 0/0 | 2/2 | 0/0 | 1/3 | 0/0 | 1/2 | 0/2 | 2/3 |

**Table 3:** Accumulative success rates of robot-plant interaction in remote leaf disease detection

| Experiment | Total diseased spots | Detected diseased spots | Success rate | Diseased plants | Identified diseased plants | Success rate in identifying diseased plants |
|---|---|---|---|---|---|---|
| Disease_Insp_1 | 20 | 10 | 50% | 8 | 6 | 75% |
| Disease_Insp_2 | 20 | 12 | 60% | 8 | 7 | 88% |

Table 2 presents the success of robot-plant interaction by comparing the number of diseased spots seen by the user (numerator) to the actual number of diseased spots determined by a human expert (denominator). The summary of these statistics is provided in Table 3. In the first experiment (Disease_Insp_1), the robotic system correctly identified all diseased spots in 2 out of 14 plants (plants 1 and 8), while partially identifying diseased spots in 4 plants (plants 3, 10, 12, and 14). For 2 plants with a diseased spot confirmed present by the human expert (plants 6 and 13), the system failed to identify any diseased spots. For the remaining 6 plants (plants 2, 4, 5, 7, 9, and 11), no diseased spots were present according to the human expert's assessment, so no detection was possible or required.

In the second experiment (Disease_Insp_2), there was a slight improvement in the robotic system's performance. The system correctly identified all diseased spots in 3 out of 14 plants (plants 1, 6, and 8), an increase from 2 plants in the first experiment, as plant 6 improved from zero detection (0 of 1 spot) to full detection (1 of 1 spot). Diseased spots continued to be partially identified in 4 plants (plants 3, 10, 12, and 14); of these, plant 14 improved from 1 of 3 spots detected to 2 of 3. Plant 13, which had 2 confirmed diseased spots, remained the sole case where no diseased spots were detected in either experiment, indicating a persistent detection failure for this specific plant that did not improve with operator experience.

Overall, while the robotic system showed consistent performance across both experiments, there were incremental improvements in its ability to detect diseased spots, particularly for plants it initially failed to identify correctly. Increasing familiarity of the human operator with the system and its controls is one plausible explanation for this pattern, as more experienced operators may guide the robotic arm more effectively; however, given the small number of diseased spots per plant (Table 2), this improvement was not formally tested for statistical significance and should be treated as an observed trend rather than a confirmed effect.

## 3.2 Soil Moisture Level Assessment

The result in this section presents the outcome of the soil moisture assessment experiment. The camera feeds obtained for the gripper camera during this experiment is represented in Figure 13. This image captures the moisture evaluation process of a Giant taro plant.

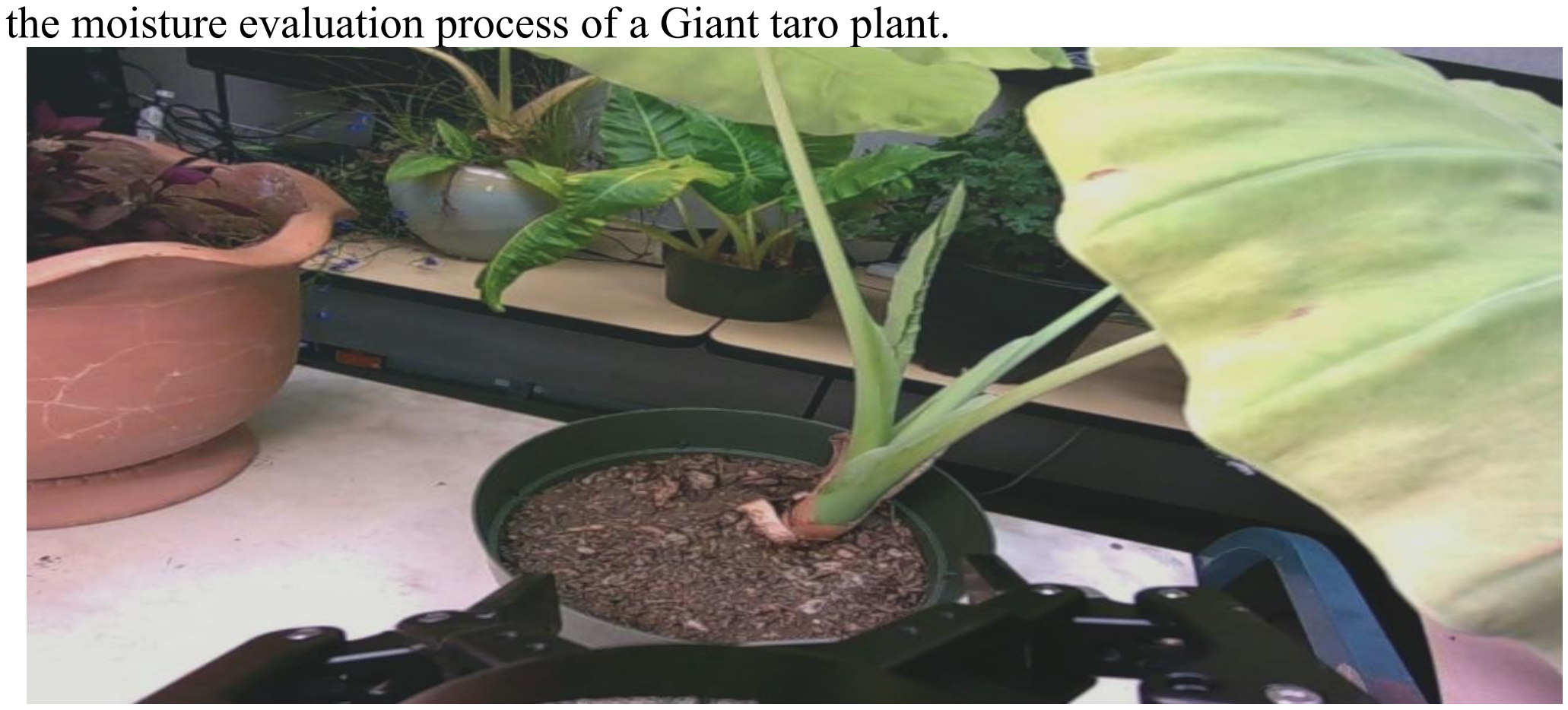

**Figure 13:** An image of a Giant Taro plant captured the gripper camera during the soil moisture assessment experiment.

The cycle completion time for the soil moisture assessment experiment is presented in Figure 14 and a summary of these statistics is given in Table 4; and the success rate of the robot plant interaction for this experiment is given in Table 5.

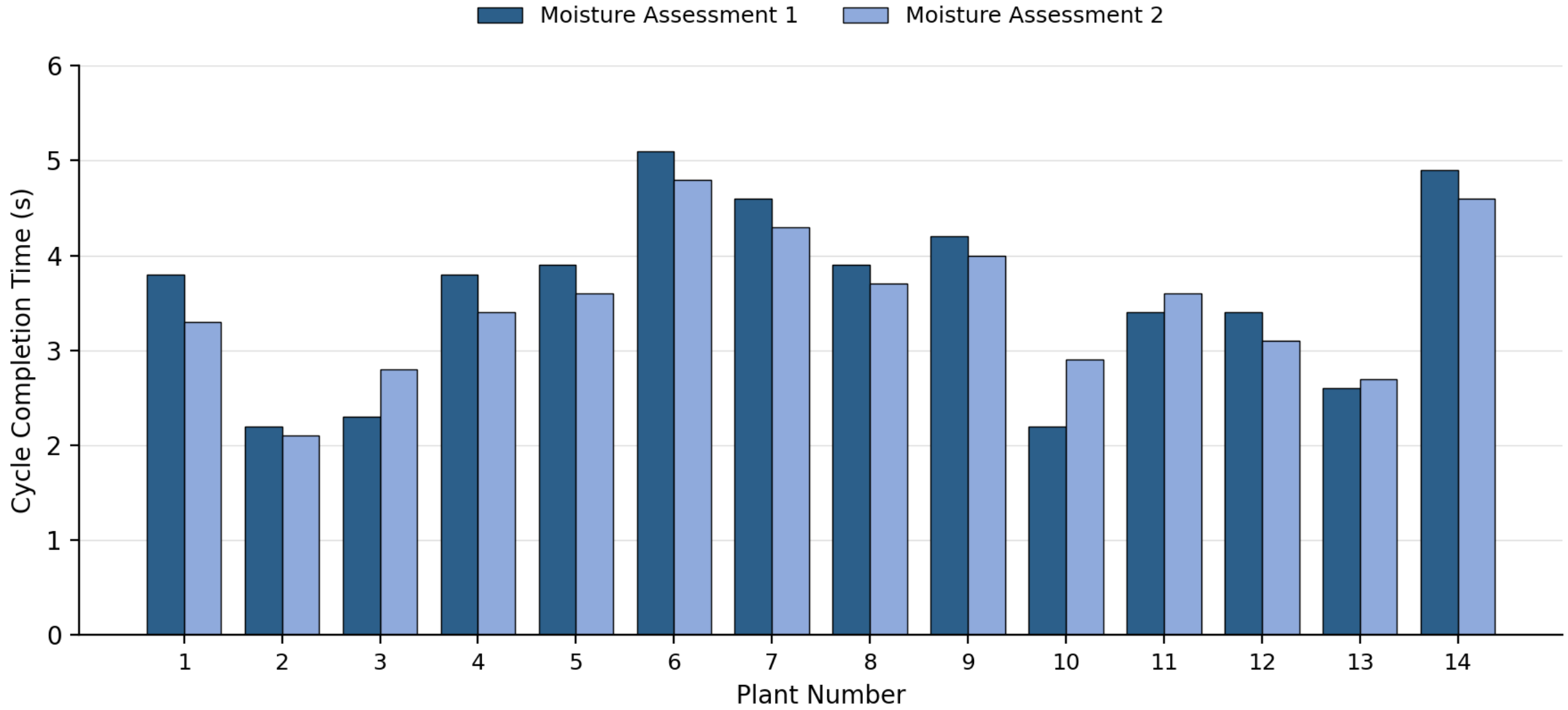

**Figure 14:** Completion times across the moisture level observation experiments

**Table 4.** Descriptive statistics of the CCT for the moisture level observation experiments across 14 heterogenous plants

| Experiment | # of assessed plants | Min (s) | Max (s) | Mean (s) | STD (s) |
|---|---|---|---|---|---|
| Moist_Insp_1 | 14 | 2.2 | 5.1 | 3.6 | 1.0 |
| Moist_Insp_2 | 14 | 2.1 | 4.8 | 3.5 | 0.7 |

The Table 4 presents data from the soil moisture assessment experiment using the developed robotic system comprising the unmanned ground platform and a robotic manipulator, where 14 distinct plants were inspected to determine their watering status across two experiments using the VR teleoperation. In the first experiment, the cycle completion times varied between 2.2 s to 5.1 s. The fastest inspections were observed for the second and tenth plants, each taking 2.2 s. In contrast, the sixth plant took the longest time at 5.1 s, followed by the seventh and fourteenth plants, which took 4.6 and 4.9 s, respectively. The shorter inspection times for the second and tenth plants can be attributed to their simpler structures, with fewer leaves or branches, making it easier and quicker for the robotic system to navigate and assess their soil moisture. Additionally, the soil moisture levels of these plants were within a normal range, requiring less time for the operator to make an accurate assessment compared to plants with more variable moisture levels. Most of the cycle times clustered around the 3-4 second range, indicating a relatively consistent performance across most plants, but a few outliers extended the completion times.

In the second experiment, there was a slight numerical decrease in mean cycle completion time. The times for most plants decreased slightly, such as the first plant (from 3.8 to 3.3 s) and the sixth plant (from 5.1 to 4.8 s). The fastest times were still observed for the second plant (2.1 s) and the third plant showed a slight increase in the assessment completion time (from 2.3 to 2.8 s). Some plants, like the tenth and eleventh, showed increases in their inspection times rather than decreases, indicating variability across plants. A paired t-test comparing CCT across all 14 plants between the two experiments confirmed that this overall reduction was not statistically significant ($t=1.07$, $p=0.306$). As detailed in Section 3.3.1, this variability is better explained by plant canopy morphology than by a general trend of operator improvement across trials.

**Table 5:** Success rates of robot-plant interaction across the moisture level assessment experiments

| Experiment | 1 | 2 | 3 | 4 | 5 | 6 | 7 | 8 | 9 | 10 | 11 | 12 | 13 | 14 |
|---|---|---|---|---|---|---|---|---|---|---|---|---|---|---|
| Moist_Insp_1 | Pass | Pass | Pass | Fail | Fail | Fail | Fail | Pass | Pass | Pass | Fail | Fail | Pass | Fail |
| Moist_Insp_2 | Pass | Pass | Pass | Fail | Fail | Fail | Pass | Pass | Pass | Pass | Fail | Fail | Pass | Pass |

As shown in Table 5, in the first experiment for the remote moisture level assessment, the success of the robot-plant interaction varied. Out of the 14 plants, the VR user successfully determined the watering needs for seven plants (Plants 1, 2, 3, 8, 9, 10, 13). However, there were significant challenges, as seven plants (Plants 4, 5, 6, 7, 11, 12, 14) resulted in fails, indicating the VR user was unable to accurately assess the soil moisture levels compared to a human expert. These challenges included complex plant structures that made it difficult for the robotic manipulator to navigate and assess soil moisture accurately, sensor limitations that affected the detection of moisture levels, and environmental factors such as variations in lighting and temperature within the greenhouse. Additionally, familiarity with the system impacted the ability to control the robotic manipulator effectively and interpret sensor data correctly. Potential technical glitches in the VR teleoperation system and soil variability in terms of composition and texture also contributed to inconsistent readings. These factors collectively led to difficulties in accurately determining the watering needs for the seven plants. This variability suggests initial limitations in the robotic system's ability to consistently determine watering needs across different plants.

In the second experiment, the success rate showed noticeable improvement. The VR user achieved successful moisture level assessment of nearly 65%. The operator successfully determined the water status for nine plants (Plants 1, 2, 3, 7, 8, 9, 10, 13, 14), while the number of fails decreased to five plants (Plants 4, 5, 6, 11, 12). This numerical increase was not statistically significant (McNemar's test, $p=0.50$, Section 3.3.2), so it should not be attributed with confidence to operator familiarity or system improvement between trials. The consistent pass results for certain plants (1, 2, 3, 8, 9, 10, 13) across both experiments highlight the system's reliability for those specific plants, while the persistent fails for others (4, 5, 6, 11, 12) suggest that further refinement is needed to enhance detection accuracy for a subset of plants; as shown in Section 3.3, this split is explained by canopy morphology rather than operator learning. Overall, the data underscored the progress in the system's performance and highlighted areas needing further optimization for reliable soil moisture assessment. These optimizations could include improving the camera quality, enhancing the skill level of the user, and improving the lighting in the environment. Getting reliable and effectives results for this experiment allow us to take into consideration integrating a water talk to the platform to carry out watering of the pots without a need for a human to be present.

## 3.3 Canopy Morphology and Occlusion-Driven Failure

The plant-level results in Sections 3.1 and 3.2 suggest that failure was not distributed randomly across the 14 plants, but instead tracked a specific physical property of the plant: canopy morphology. Plants were classified post hoc into two morphology classes based on leaf structure: broad/single-leaf plants (zonal geranium, taro, giant taro, flaming flower, Gollum Jade, and konjac), whose foliage presents a small number of large, largely unobstructed leaf and soil surfaces to the gripper camera, and dense/compound plants (rose geranium, rose, and common coleus), whose foliage consists of numerous small, overlapping leaves and stems that visually and physically obstruct the camera's line of sight to the soil surface and lower leaves. Table 6 presents the watering-need assessment pass rate for each morphology class across both experiments.

**Table 6.** Soil moisture assessment pass rate by plant canopy morphology class across both experiments

| Morphology class | Plant IDs | n | Exp. 1 pass rate | Exp. 2 pass rate |
|---|---|---|---|---|
| Broad/single-leaf | 1, 2, 3, 7, 8, 9, 10, 13 | 8 | 87.5% (7/8) | 100% (8/8) |
| Dense/compound | 4, 5, 6, 11, 12, 14 | 6 | 0% (0/6) | 16.7% (1/6) |

As shown in Table 6, canopy morphology class is a strong predictor of soil moisture assessment outcome. Every broad/single-leaf plant (8 of 8) was correctly assessed by Experiment 2, and 7 of 8 were already correct in Experiment 1. In contrast, no dense/compound plant (0 of 6) was correctly assessed in Experiment 1, and only 1 of 6 improved to a correct assessment by Experiment 2. A Fisher's exact test confirms that this association between morphology class and assessment outcome is statistically significant in both experiments (Experiment 1: $p=0.0047$; Experiment 2: $p=0.0030$), rather than a pattern arising from the small sample size. This pattern is consistent with an occlusion-driven failure mechanism: dense, overlapping foliage physically blocks the gripper camera's view of the soil surface, preventing the operator from forming an accurate moisture judgment regardless of operator skill or system familiarity, whereas broad, single leaves leave the soil surface visually accessible from most approach angles. Figure 15 illustrates this split.

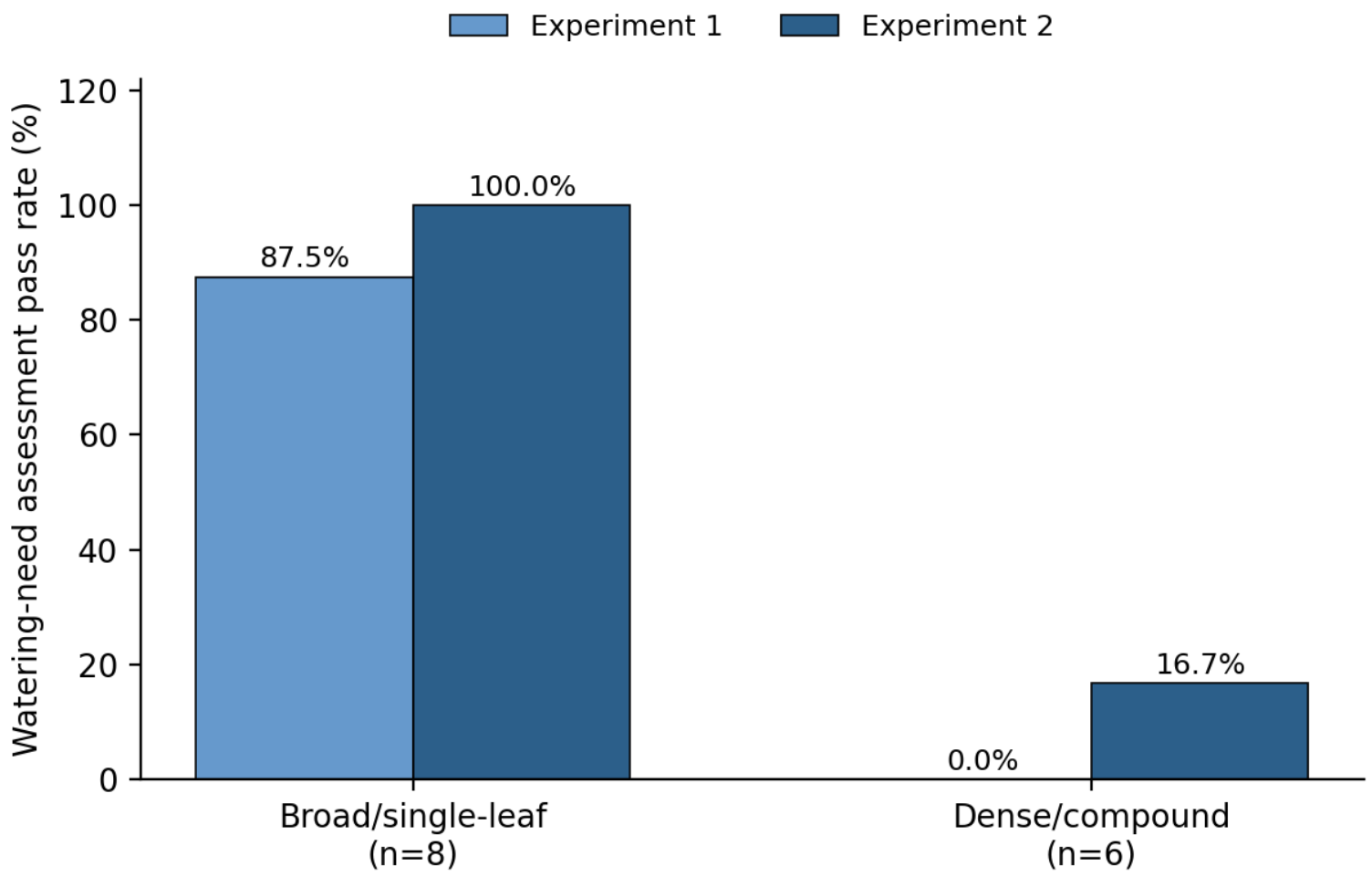

**Figure 15:** Soil moisture assessment pass rate by canopy morphology class, Experiment 1 vs. Experiment 2

Notably, this morphology effect was specific to the soil moisture assessment task and did not hold as cleanly for leaf disease detection (Section 3.1): plant 13 (giant taro, broad/single) had zero diseased spots detected despite its favorable morphology, while dense-canopy plants 6, 12, and 14 each had at least partial success. This suggests that disease-spot detection depends more on the location of the diseased spot on the leaf surface and on lighting conditions at the time of inspection than on overall canopy density, whereas soil moisture assessment depends on maintaining an unobstructed camera view of the soil itself, a condition that dense foliage directly and consistently defeats. Taken together, these findings indicate that the dominant limiting factor for remote soil moisture assessment in this system is not operator skill or VR interface latency, but the sensor's inability to achieve a clear line of sight to the target surface through dense foliage. This points to a concrete engineering design implication for future greenhouse HRI systems: gripper-camera viewpoint strategy, and potentially camera placement or auxiliary sensing (e.g., a soil-moisture probe mounted on the gripper itself rather than a purely visual assessment), should be adapted to the canopy density of the target plant rather than treated as a fixed configuration across all plant types.

### *3.3.1 Speed-Accuracy Decoupling Under Occlusion*

If soil moisture assessment failures on dense-canopy plants were primarily due to insufficient operator time or effort, mean cycle completion time (CCT) would be expected to increase for these plants as the operator worked harder to compensate. Table 7 tests this by comparing mean CCT for the soil moisture assessment task across the two morphology classes and both experiments.

**Table 7.** Mean cycle completion time (CCT) for the soil moisture assessment task by canopy morphology class

| Morphology class | n | CCT Exp. 1 mean (s) | CCT Exp. 2 mean (s) | Δ CCT |
|---|---|---|---|---|
| Broad/single leaf | 8 | 3.23 | 3.23 | 0.0% |
| Dense/compound | 6 | 4.08 | 3.85 | −5.7% |

As shown in Table 7, mean CCT for broad/single-leaf plants was unchanged between experiments (3.23 s in both), despite the pass rate for this group rising from 87.5% to 100% (Table 6); the operator reached perfect success without needing additional time. For dense/compound plants, mean CCT decreased by 5.7% between experiments (from 4.08 s to 3.85 s), yet the pass rate for this group improved only marginally (0% to 16.7%). In other words, the operator became faster at attempting the dense-canopy plants without becoming more successful at them. This decoupling of speed from accuracy indicates that occlusion imposes a hard sensing limit that cannot be overcome by additional operator time, effort, or dwell time on the target; the constraint is the availability of visual information, not the speed or attentiveness of the human operator. This has a direct implication for teleoperation system design: operator training or interface changes intended to encourage more careful, deliberate inspection will not resolve occlusion-driven failures, and design effort is better directed toward the sensing configuration itself, for example an adjustable or multi-angle gripper camera, or a non-visual soil moisture sensor mounted on the gripper.

### *3.3.2 Summary of Hypothesis Testing*

**Table 8.** Summary of hypothesis testing outcomes

| Hyp. | Statistical test and result | Outcome | Status |
|---|---|---|---|
| H1 | Descriptive: CCT and success rate were measurable and repeatable across both tasks (Tables 1–5) | Metrics obtained as expected | Supported |
| H2 | Fisher's exact test, morphology vs. moisture pass/fail: $p=0.0047$ (Exp. 1), $p=0.0030$ (Exp. 2) | Significant association | Supported |
| H3 | Paired t-test, CCT (disease): $p=0.378$; CCT (moisture): $p=0.306$; McNemar’s test, moisture pass/fail: $p=0.50$ | No significant difference | Not supported |
| H4 | Descriptive: dense-canopy CCT fell 5.7% (Table 7) while pass rate rose only 16.7 pts (Table 6) | Speed and success decoupled | Supported |

Table 8 summarizes the outcome of formal hypothesis testing across this study. H1 and H4 are supported by the data presented in Sections 3.1 through 3.3.1. H2 is supported and statistically confirmed. Notably,

H3 (that operator experience across trials would improve performance) is not supported: neither cycle completion time nor pass/fail success rate improved significantly between the first and second experiments for either task. Improvements described in earlier sections as arising from operator familiarity should therefore be read as descriptive, non-significant trends. This null result is itself informative: it suggests that, within the range of experience gained over two trials with 14 plants, canopy morphology is a stronger determinant of task outcome than operator practice, reinforcing occlusion, rather than operator skill, as the primary constraint on system performance.

# 4. CONCLUSION

This study explored the efficacy of human-robot interaction (HRI) for leaf inspection and soil moisture assessment in a greenhouse environment. The remote interaction between an operator and robotic system took place utilizing VR technologies. Across two experiments, 14 distinct plants were inspected using VR teleoperation to assess the accuracy and efficiency of the developed system for HRI.

The results from the plant disease inspection experiments showed promising results for the robotic system, a mean and standard deviation of 6 s and 1.2 s, respectively. The experiments revealed a range of cycle completion times, with some plants taking significantly longer to inspect, 3.3 to 8 s, though this variation was not statistically significant (paired t-test, $p=0.378$). There were also variations in the accuracy of disease detection. Although the detection of diseased spots and plant-based disease achieved up to 60% and 88%, respectively, and these rates improved numerically in the second experiment, this improvement was not formally tested for significance given the small number of diseased spots per plant, so it is reported descriptively rather than as confirmed evidence of a user-experience effect. In the soil moisture assessment experiments, the cycle completion times varied from 2 to 5 s. The results of initial assessment accuracy showed a mix of successful and failed determinations of watering needs. The first experiment had a higher number of fails, particularly for plants with more complex soil moisture conditions. The second experiment showed a numerical increase in the success rate, to 64.3% (9 of 14); however, a McNemar's test on the paired pass/fail outcomes found this change not to be statistically significant ($p=0.50$), so it should not be interpreted as confirmed evidence of operator learning across trials. Indeed, this improvement was not uniform across plants: consistent success across both experiments was observed for 7 of 14 plants, while 5 of 14 plants failed consistently in both experiments, and the remaining 2 plants improved from failure to success between experiments. As detailed in Section 3.3, this split closely tracked plant canopy morphology rather than operator learning: all plants with broad, single-leaf canopies achieved a correct assessment by the second experiment, while dense, compound-canopy plants achieved only a 16.7% success rate, a difference confirmed as statistically significant by Fisher's exact test ($p=0.003$), indicating that camera occlusion by foliage, rather than operator skill, was the dominant factor limiting soil moisture assessment reliability.

Despite these promising findings, several limitations were encountered in this study. Occlusion by dense plant canopies emerged as the dominant limiting factor for soil moisture assessment, as detailed in Section 3.3, indicating that future gripper-camera viewpoint strategies, and potentially auxiliary sensing modalities, should be adapted to canopy density rather than treated as fixed across plant types. Although the adaptability of the developed robotic system to different plant conditions (type, size, and shape) is achieved, the variability in inspection times and detection accuracy indicated potential limitations in the VR teleoperation setup, and the single-operator design of this study means that improvements between experiments cannot be fully attributed to system performance independent of operator learning. Additionally, the canopy morphology classification used in Section 3.3 was performed post hoc, after outcomes were known, which carries some risk of classification bias; future studies should pre-register morphology criteria prior to data collection to rule this out. In future work, the focus will be on refining the robotic system's sensing strategy to overcome occlusion issues for dense-canopy plants, enhancing VR teleoperation precision, and improving the overall robustness of the system for diverse greenhouse environments.

## DECLARATION OF COMPETING INTERESTS

The authors declare that they have no known competing financial interests or personal relationships that could have appeared to influence the work reported in this paper.

## DATA AND CODE AVAILABILITY:

The data used in this study is available from the corresponding author upon reasonable request.

## CREDIT AUTHORSHIP CONTRIBUTION STATEMENT

**Daniel Udekwe:** Writing – review & editing, Writing – original draft, Validation, Methodology, Investigation, Formal analysis, Data curation.
**Hassan Seyyedhasani:** Funding acquisition, Methodology, Resources, Writing-review & editing, Project administration

## ACKNOWLEDGEMENT

This work is partially supported by the USDA National Institute of Food and Agriculture (NIFA) Hatch Multistate project under VA-160172, 2022. Primary Critical Issue: SmartFarm Technology and Security